\documentclass[PHME]{PHMSociety}

\PassOptionsToPackage{hyphens}{url}
\usepackage{xurl}

\usepackage{apacite}
\makeatletter
\providecommand\@xviiipt{17.28}
\providecommand\@xxiiipt{22.00}
\makeatother
\usepackage{enumitem}
\usepackage{cuted}
\usepackage{graphicx}
\usepackage{amsmath,amssymb}
\usepackage{booktabs}
\usepackage{tabularx}
\usepackage{ragged2e}
\usepackage{float}
\usepackage{fontawesome5}
\usepackage{subcaption}
\usepackage{caption}
\usepackage{svg}
\usepackage{comment}
\usepackage{algpseudocode}
\usepackage{algorithm2e}
\usepackage[most]{tcolorbox}

\tcbset{
  colback=black!2,
  colframe=black!35,
  boxrule=0.5pt,
  arc=2mm,
  left=2.5mm,right=2.5mm,top=1.5mm,bottom=1.5mm,
  enhanced, breakable
}
\newtcolorbox{promptbox}[1]{title={#1}, fonttitle=\bfseries, parbox=false}

\begin{document}

\title{Time‑Series Retrieval for Grounding Multimodal Language Models in Remaining Useful Life Prediction}

\author{%
	Valeriu Dimidov and Raphaël Frank
}

\address{
	\affiliation{}{Interdisciplinary Centre for Security, Reliability and Trust (SnT) \\
    University of Luxembourg \\
    29 Avenue J.F. Kennedy L-1855, Luxembourg}{
		{\email{firstname.lastname@uni.lu}}
		}
}

\maketitle
\pagestyle{fancy}
\thispagestyle{plain}

\phmLicenseFootnote{FirstAuthorFirstName FirstAuthorLastName}

\begin{abstract}
Large language models (LLMs) and agentic AI systems are increasingly being explored for domain-specific maintenance and prognostics tasks, raising the question of whether they can effectively support prognostics and health management (PHM). In this paper, we investigate remaining useful life (RUL) estimation with multimodal large language models (MLLMs) grounded through time-series retrieval. We propose a framework in which historically similar degradation segments are retrieved from the training set and, together with the test trajectory, transformed into a visual comparison artifact that is processed by the MLLM through a structured multimodal prompt. The approach is evaluated on the FD001 partition of the C-MAPSS benchmark under repeated experiments comparing retrieval-based inference against a non-retrieval baseline based on random reference selection. The results show that time-series retrieval consistently improves MLLM-based RUL prediction across the evaluated models, yielding lower error and more stable performance. At the same time, the magnitude of the benefit depends on model capacity, indicating that retrieval is most effective when the underlying MLLM is able to exploit the retrieved evidence. Overall, the study shows that time-series RAG is a promising mechanism for improving multimodal prognostic reasoning, while also highlighting the current limitations of MLLM-based RUL estimation in practical PHM settings.
\end{abstract}

\section{Introduction}
\label{sec:intr}

Predictive Maintenance (PdM) aims to reduce downtime, maintenance costs, and operational risks by anticipating failures before they occur. A key task in this context is Remaining Useful Life (RUL) estimation, which provides an estimate of the time left before a component reaches failure. Accurate RUL estimation supports maintenance planning, resource allocation, and risk-aware decision making in Prognostics and Health Management (PHM).

Conventional RUL estimation methods are typically based on statistical degradation models, physics-informed approaches, or data-driven architectures trained on sensor time series. Although these methods can achieve strong predictive performance, they usually operate only on numerical data and provide limited support for reasoning over external contextual knowledge.

Recent advances in large language models (LLMs) and multimodal large language models (MLLMs) create an opportunity to revisit RUL estimation from a different perspective \cite{zhangLargeLanguageModels2024,yinSurveyMultimodalLarge2024}. Instead of relying only on a learned numerical predictor, an MLLM can be prompted with visual and textual evidence describing the target trajectory. Retrieval-augmented generation (RAG) can further enrich this input by selecting historical degradation segments that are similar to the query trajectory, following the broader idea of grounding language-model generation with retrieved external evidence \cite{lewisRetrievalAugmentedGenerationKnowledgeIntensive2021}. However, in the RUL setting, the effect of such retrieval is not obvious: relevant references may ground the model on useful degradation evidence, whereas poorly matched references may introduce noise or bias the prediction.

Therefore, this paper studies whether time-series retrieval improves MLLM-based RUL estimation. We propose a framework in which similar degradation segments are retrieved from historical run-to-failure trajectories and combined with the query trajectory into a visual comparison artifact. The same multimodal prompt structure is then used to compare retrieval-based inference with a non-retrieval baseline based on random reference selection on the FD001 partition of the C-MAPSS benchmark \cite{saxenaDamagePropagationModeling2008c,decastroDeanFrederickSaratoga2007}.

The main contributions of this study are:
\begin{itemize}
    \item A multimodal framework that reformulates RUL estimation as an evidence-grounded visual reasoning task for MLLMs.
    \item A time-series RAG mechanism that retrieves historical degradation segments and presents them to the MLLM as trajectory-level reference evidence.
    \item A controlled experimental comparison between retrieval-based and random-reference inference on the FD001 partition of the C-MAPSS benchmark.
    \item An analysis of the benefits, limitations, and failure modes associated with retrieval augmentation in the RUL estimation setting.
\end{itemize}

The remainder of the paper is organized as follows: Section~\ref{sec:rel} reviews related work, Section~\ref{sec:met} presents the methodology, Section~\ref{sec:res} reports the experimental results, Section~\ref{sec:disc} discusses the main findings and limitations, and Section~\ref{sec:conc} concludes the paper.

\begin{figure*}
    \centering
    \includegraphics[width=1\linewidth]{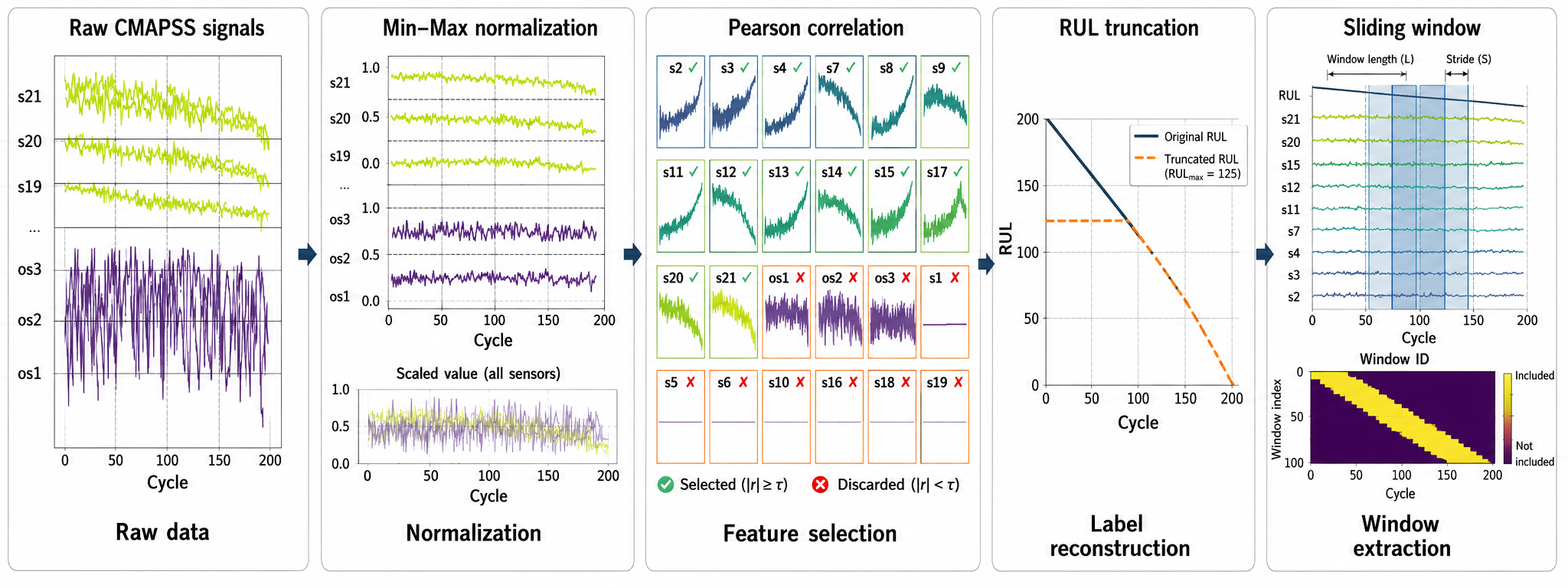}
    \caption{Preprocessing workflow applied to the FD001 partition of the C-MAPSS dataset.}
    \label{fig:data_pipeline}
\end{figure*}

\begin{table*}[t]
\centering
\caption{Overview of the main pipeline modules according to their inputs, transformations, and outputs.}
\label{tab:transformation_modules}
\small
\renewcommand{\arraystretch}{1.15}
\begin{tabularx}{\textwidth}{
p{0.22\textwidth}
>{\RaggedRight\arraybackslash}p{0.25\textwidth}
>{\RaggedRight\arraybackslash}X
>{\RaggedRight\arraybackslash}p{0.22\textwidth}}
\toprule
\textbf{Module} & \textbf{Input} & \textbf{Transformation} & \textbf{Output} \\
\midrule

Segment encoder &
Normalized sliding-window segment &
Encode with an LSTM autoencoder. &
Fixed-dimensional embedding. \\

Retrieval memory builder &
Training segments and RUL labels &
Sample by RUL bin and index embeddings. &
Balanced vector database. \\

Query retriever &
Last observed test segment &
Encode query and perform $k$-nearest-neighbor search. &
Top-$k$ historical references. \\

Trajectory comparison builder &
Query trajectory and retrieved references &
Align trajectories and generate a multisensor plot. &
Trajectory-comparison image. \\

Prompt composer &
Comparison image and task instructions &
Insert visual evidence into a structured prompt. &
Multimodal prompt for RUL estimation. \\

\bottomrule
\end{tabularx}
\end{table*}

\section{Related Work}
\label{sec:rel}

LLMs have started to attract attention in time-series analysis. Recent survey studies indicate that this line of research mainly follows several directions, including direct prompting, time-series quantization into token-like representations, alignment of numerical sequences with language-model spaces, the use of visual representations as an intermediate reasoning interface, and the integration of LLMs with external tools \cite{zhangLargeLanguageModels2024}. In parallel, the rapid development of multimodal large language models (MLLMs) has shown that language-centered models can reason jointly over heterogeneous modalities such as text, images and audio, which is particularly relevant when sensor trajectories are converted into plots or textual summaries before inference \cite{yinSurveyMultimodalLarge2024}.

In the specific context of PHM and RUL estimation, the use of LLMs and MLLMs is still recent but growing. One contribution introduced an LLM-based regression framework for turbofan RUL prediction and reported competitive results together with promising transfer-learning behavior \cite{chenRemainingUsefulLife}. Another study explored a pre-trained LLM-based approach for aircraft-engine RUL prediction on C-MAPSS \cite{tanPRETRAINEDLLMBASEDREMAINING2025}. More recently, a multimodal framework was proposed to jointly exploit temporal signals, frequency-domain images, and textual domain knowledge for industrial time-series analysis, and it was evaluated on the four standard C-MAPSS subsets \cite{wangTSMLLMMultiModalLarge2026}. At a broader PHM level, another work proposed a language-model-based framework intended to support multiple maintenance-related tasks within a unified setting \cite{renPHMGPTLarge2025}.

Another relevant direction concerns the integration of external knowledge into LLM-based PHM pipelines. RAG was introduced to combine parametric language models with explicit non-parametric memory, thereby improving grounding and updateability \cite{lewisRetrievalAugmentedGenerationKnowledgeIntensive2021}. In predictive maintenance, recent studies have argued that retrieval and knowledge augmentation can help LLM-based systems access maintenance records, technical documentation, and operational procedures more effectively \cite{jiangArtificialIntelligenceAgentEnabled2025}. 

Despite these advances, only a limited number of studies explicitly investigate the influence of retrieval augmentation on MLLM-based RUL estimation in a controlled setting. Existing studies mainly focus either on adapting LLMs and MLLMs to prognostics tasks or on using textual retrieval for broader PHM support. Consequently, the effect of time-series RAG on RUL prediction quality remains insufficiently studied \cite{chenRemainingUsefulLife,tanPRETRAINEDLLMBASEDREMAINING2025,wangTSMLLMMultiModalLarge2026,jiangArtificialIntelligenceAgentEnabled2025,hafsiPotentialGenerativeArtificial2025,kirubanandanCausalAwareLLMAgents2025}.

\section{Methodology}
\label{sec:met}

This section describes the problem setup, proposed model pipeline, and experimental configuration.

\subsection{Problem Definition}
\label{sec:problem_definition}

We consider a set of $n$ equipment units, denoted by
$U = \{U_{0}, U_{1}, \dots, U_{n-1}\}$, monitored over time through onboard measurements. Each unit $U_i$ is represented by a multivariate time series
$X_i \in \mathbb{R}^{T_i \times M}$, where $T_i \in \mathbb{N}$ denotes the number of observed cycles and $M$ is the number of monitored variables. The corresponding sequence of RUL targets is denoted by $R_i \in \mathbb{N}_0^{T_i}$.

Given the dataset
\begin{equation}
    \mathcal{D} = \{(X_i, R_i)\}_{i=0}^{n-1},
\end{equation}
the objective is to learn a mapping function
\begin{equation}
\label{eq:mapping_function}
    f_{TS}: \bigcup_{T \in \mathbb{N}} \mathbb{R}^{T \times M} \rightarrow \mathbb{N}_0
\end{equation}
such that, for a given time series $X_i$, the function $f_{TS}$ predicts the RUL of its last observation.

In this work, the mapping function $f_{TS}$ is approximated using a multimodal large language model.
Since MLLMs are designed to process multimodal inputs such as images and text, each time series $X_i$ is transformed into a multimodal representation $Z_i$, which may include visual depictions of sensor trajectories and textual contextual information.
The prediction function can therefore be expressed as
\begin{equation}
    f_{MM} : \mathcal{Z} \rightarrow \mathbb{N}_0,
\end{equation}
where $\mathcal{Z}$ denotes the space of multimodal representations derived from the original time series.
The goal remains the estimation of the RUL of the last observation of each unit.

\subsection{Dataset}
\label{sec:dat}

This study uses the C-MAPSS benchmark for the empirical evaluation of RUL estimation. C-MAPSS is a widely used simulated turbofan-engine degradation dataset in which each trajectory corresponds to one equipment unit monitored over successive operating cycles through three operational setting variables and multiple sensor measurements. The training split contains complete run-to-failure trajectories, whereas the test split contains truncated trajectories that end before failure. Accordingly, the task is to estimate the RUL at the last observed cycle of each test unit \cite{saxenaDamagePropagationModeling2008c,decastroDeanFrederickSaratoga2007}.

Owing to economic, computational, and time constraints, the experimental campaign reported in this paper is restricted to the FD001 partition. This design choice enables a focused and controlled analysis within a feasible experimental budget, while a broader evaluation on the remaining partitions is deferred to future work.

For further details on the simulator and dataset construction, the reader is referred to \cite{saxenaDamagePropagationModeling2008c,decastroDeanFrederickSaratoga2007}.

\begin{figure*}
    \centering

    \begin{subfigure}{0.96\linewidth}
        \centering
        \includegraphics[width=\linewidth]{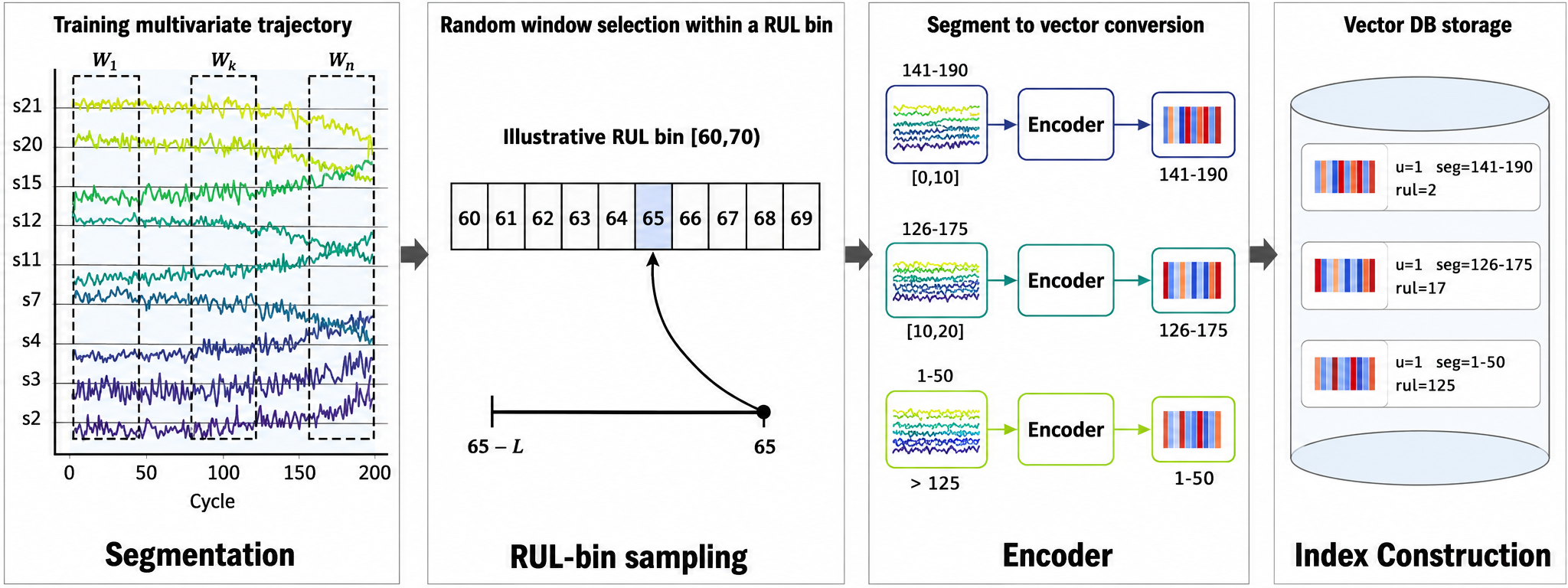}
        \caption{RAG builder. Training trajectories are segmented, sampled by RUL bin, encoded, and indexed in the vector database.}
        \label{fig:rag_offline}
    \end{subfigure}

    \vspace{0.6em}

    \begin{subfigure}{0.96\linewidth}
        \centering
        \includegraphics[width=\linewidth]{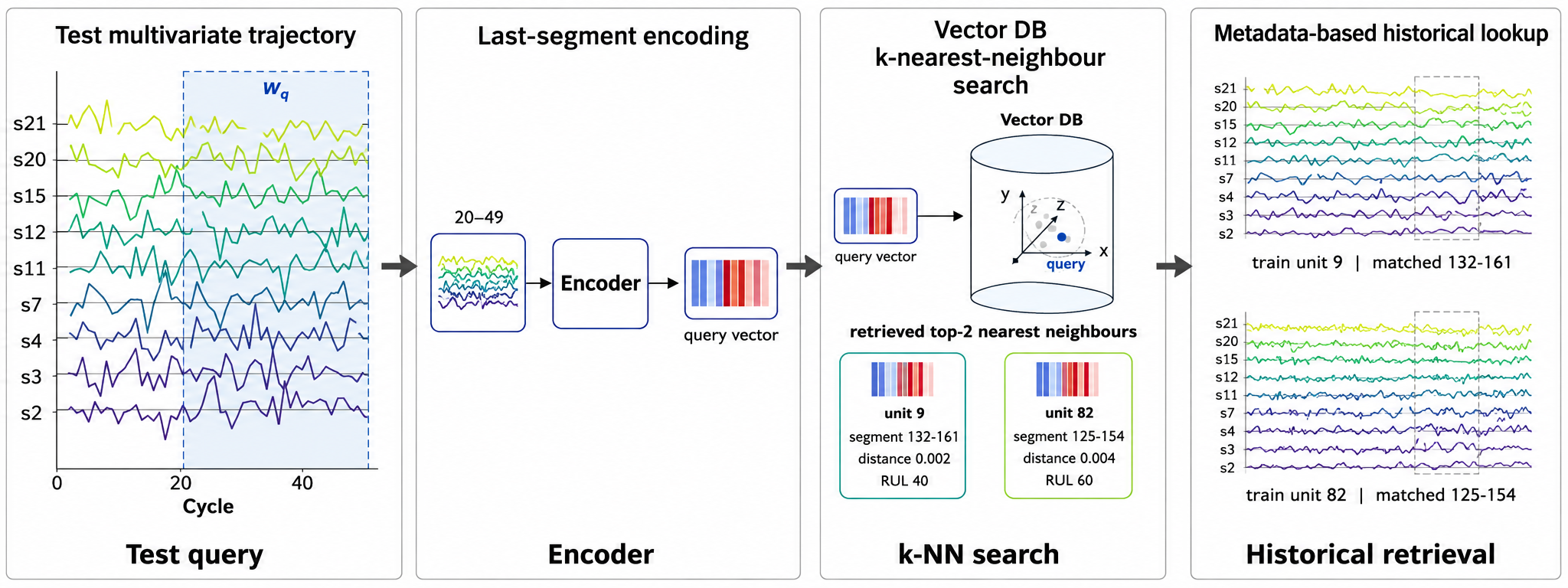}
        \caption{Query retriever. The last segment of the test trajectory is encoded and used to retrieve the nearest indexed train segments.}
        \label{fig:rag_online}
    \end{subfigure}

    \caption{Overview of the time‑series RAG mechanism: (a) offline indexing of encoded training segments; (b) online retrieval using the encoded test segment.}
    \label{fig:rag_pipeline}
\end{figure*}

\subsection{Preprocessing}
\label{sec:preprocessing}

The preprocessing pipeline transforms the raw C-MAPSS trajectories into standardized fixed-length samples suitable for both retrieval and multimodal inference, as illustrated in Figure~\ref{fig:data_pipeline}. The raw trajectories are first grouped by equipment unit and ordered by cycle. Next, min--max normalization is applied independently to all sensor channels using statistics computed from the training set. This transformation reduces scale discrepancies across variables and yields a more comparable representation of the degradation behavior across units.

After normalization, feature selection based on Pearson correlation is applied to reduce redundancy and eliminate non-informative measurements. Only the variables selected through this process are kept for the subsequent stages.

The cycle-level RUL targets are then reconstructed from the distance to failure, and a truncation threshold of 125 cycles is applied. Finally, each normalized trajectory is segmented through a sliding-window procedure. Given a window length $L$, the method extracts fixed-length temporal segments, and each segment is associated with the RUL value of its last cycle. This step converts variable-length degradation trajectories into standardized samples that can be used consistently by the retrieval module and by the multimodal inference pipeline.

\subsection{Time-Series RAG Mechanism}
\label{sec:ts_rag}

The proposed time-series RAG mechanism is composed of three main modules: the encoder module, the retrieval memory builder, and the query retriever. Figure~\ref{fig:rag_pipeline} summarizes the overall process, distinguishing between the offline construction of the retrieval memory and the online retrieval of historical references for a test unit.

\subsubsection{Encoder Module}
\label{sec:encoder_module}

The encoder module transforms each normalized sliding-window segment into a compact vector representation used for similarity-based retrieval. Its purpose is to represent the recent temporal behavior of a unit, including sensor levels and local degradation trends, in a fixed-dimensional embedding space.

Let $S_j$ denote a normalized segment produced by the preprocessing step introduced in Section~\ref{sec:preprocessing}. The encoder defines a mapping
\[
z_j = \phi(S_j) \in \mathbb{R}^{d}
\]
where $z_j$ is the embedding of segment $S_j$ and $d$ is the embedding dimension.

In this work, the encoder is obtained through an LSTM-based autoencoding scheme trained exclusively on windows extracted from the training trajectories. The LSTM encoder processes the multivariate sensor segment and maps it to a latent embedding. During training, a lightweight decoder is attached to this embedding and optimized to reconstruct the final sensor state of the segment.

After training, the decoder is discarded and only the encoder, together with the embedding projection, is retained. The main architectural and training settings of the encoder are summarized in Table~\ref{tab:lstm_encoder_configuration}.

\begin{table}[t]
\centering
\caption{Configuration of the learned LSTM encoder used to generate retrieval embeddings.}
\label{tab:lstm_encoder_configuration}
\small
\renewcommand{\arraystretch}{1.10}
\begin{tabular}{p{0.43\columnwidth} p{0.45\columnwidth}}
\toprule
\textbf{Aspect} & \textbf{Configuration} \\
\midrule
Input window length & $L = 30$ cycles \\
Input variables & Selected sensor features \\
Encoder type & LSTM encoder \\
Number of recurrent layers & 1 \\
Embedding layer & Linear projection \\
Embedding dimension & $d = 64$ \\
Hidden dimension & 128 \\
Decoder type & Feed-forward head \\
Training objective & Final-step reconstruction \\
Loss function & Mean squared error \\
Optimizer & AdamW \\
Training epochs & 200 \\
Batch size & 128 \\
Learning rate & $10^{-3}$ \\
Weight decay & $10^{-5}$ \\
Embedding normalization & L2 normalization \\
\bottomrule
\end{tabular}
\end{table}

\subsubsection{Retrieval Memory Builder}
\label{sec:retrieval_memory_builder}

The retrieval memory builder constructs the non-parametric memory used by the proposed time-series RAG mechanism. This module corresponds to the offline stage shown in Figure~\ref{fig:rag_pipeline}(a). It operates only on the training partition, in order to avoid information leakage, and transforms historical degradation data into a searchable collection of representative examples.

\paragraph{Window-to-Embedding Transformation.}
The module is built from the windowed training segments produced by the preprocessing step described in Section~\ref{sec:preprocessing}. Each segment is associated with the RUL value of its last cycle. The trained encoder described in Section~\ref{sec:encoder_module} is then used to transform each selected segment into a fixed-dimensional embedding.

\paragraph{RUL-Balanced Retrieval Memory Construction.}
To reduce redundancy and improve coverage across degradation stages, the training segments are grouped according to their RUL values, and a subset of representative segments is sampled from each RUL bin.

The resulting retrieval memory can be represented as
\[
\mathcal{C} =
\left\{
(z_j, r_j, m_j)
\right\}_{j=1}^{N}
\]
where $z_j$ is the segment embedding defined by the encoder module, $r_j$ is the associated RUL value, and $m_j$ contains metadata such as the source unit and cycle position. The embeddings and metadata are then stored in a vector database, forming a searchable memory of historical degradation examples.

\begin{figure*}[t]
    \centering
    \begin{subfigure}[t]{0.49\linewidth}
        \centering
        \includegraphics[width=\linewidth]{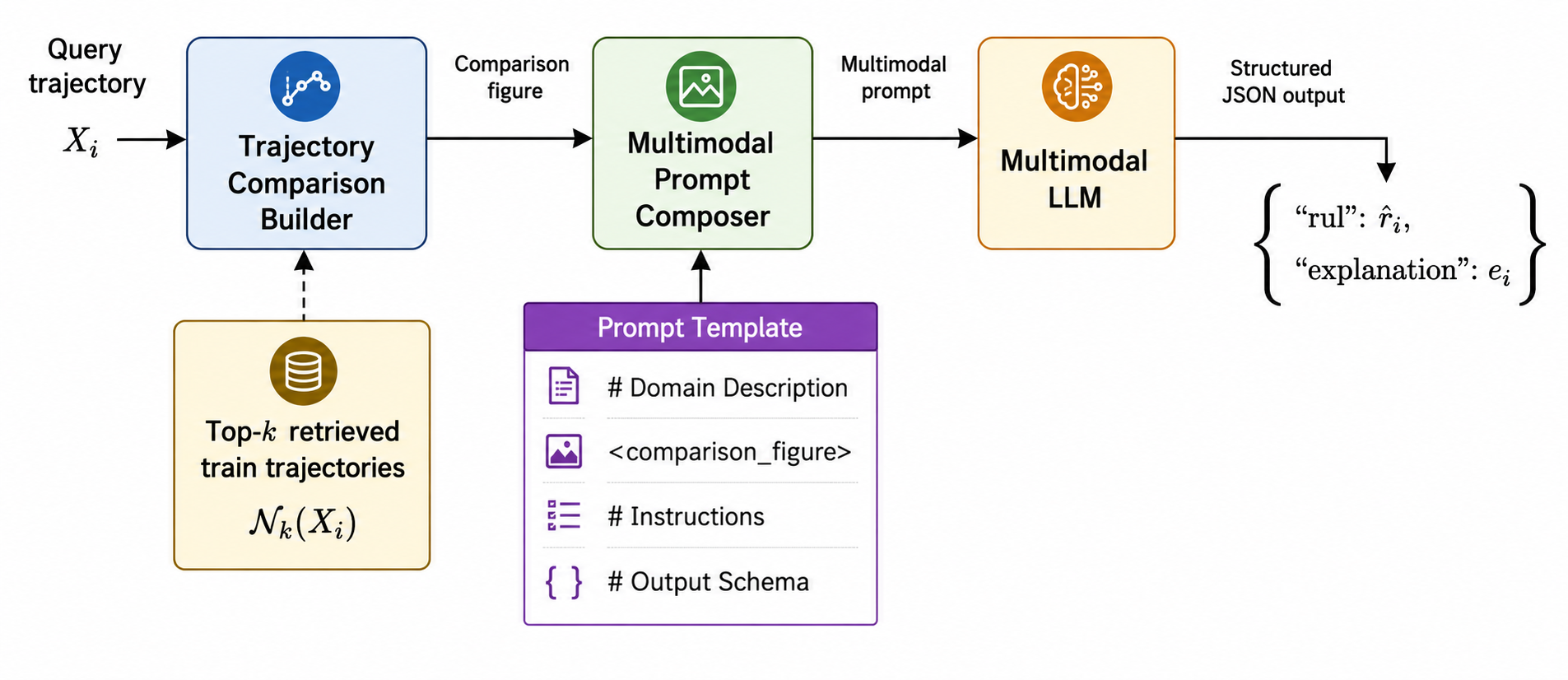}
        \caption{Multimodal prompt-generation workflow.}
        \label{fig:mm_prompt_workflow}
    \end{subfigure}
    \hfill
    \begin{subfigure}[t]{0.49\linewidth}
        \centering
        \includegraphics[width=\linewidth]{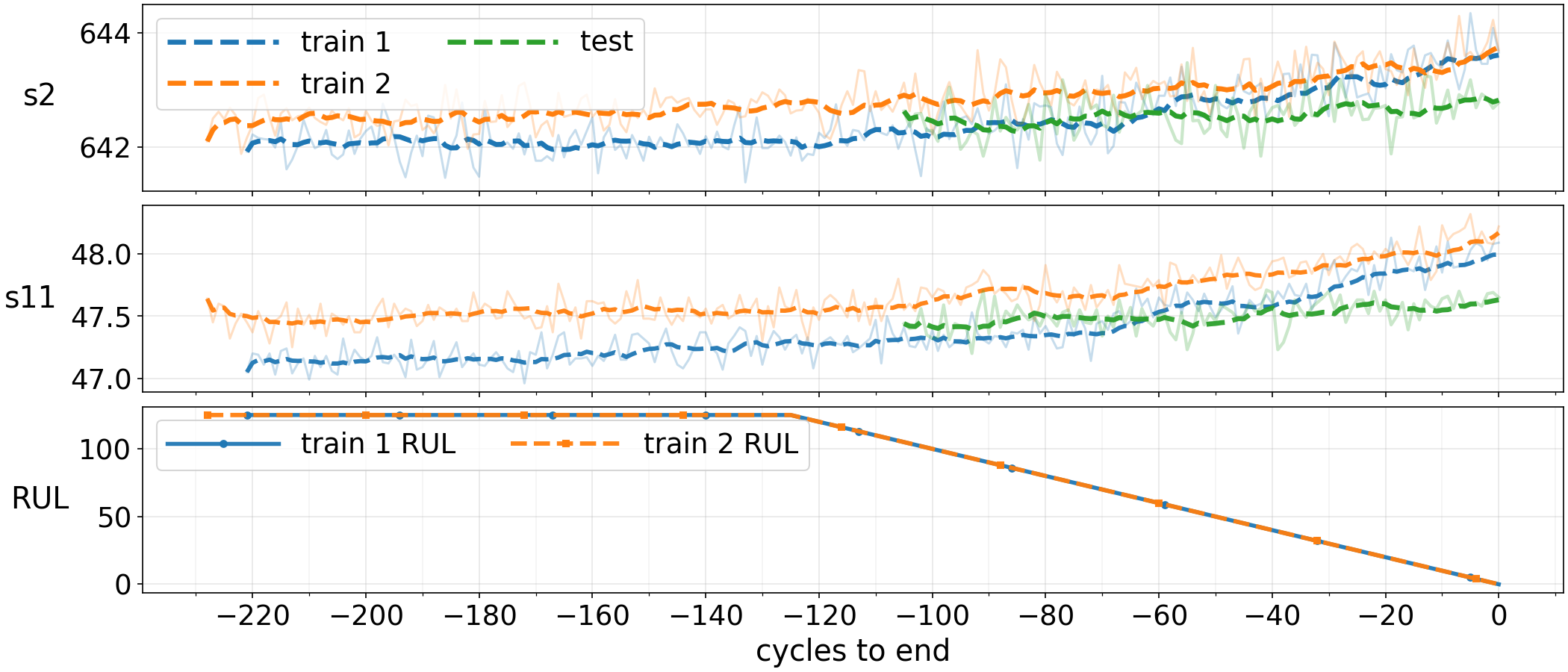}
        \caption{Example of the trajectory-comparison image.}
        \label{fig:mm_prompt_example}
    \end{subfigure}
    
    \caption{Multimodal prompt generation and example comparison artifact used for MLLM inference.}
    \label{fig:mm_prompt_generation}
\end{figure*}

\subsubsection{Query Retriever}
\label{sec:query_retriever}

The query retriever is the online module responsible for selecting the historical references used to ground the MLLM prediction. This module corresponds to the online stage shown in Figure~\ref{fig:rag_pipeline}(b). For each test unit, only the last observed segment is used as the query, since the objective is to estimate the RUL at the most recent cycle.

The query segment is encoded using the same encoder adopted during memory construction. This ensures that both training and test segments are represented in the same embedding space. Given the last observed query segment $Q_i$, its embedding is obtained as
\[
z_i^{(q)} = \phi(Q_i)
\]

A nearest-neighbor search is then performed in the retrieval memory to identify the $k$ most similar historical segments:
\[
\mathcal{N}_{k}(Q_i)
=
\operatorname{top}\text{-}k_{j \in \{1,\ldots,N\}}
\operatorname{sim}
\left(
z_i^{(q)}, z_j
\right)
\]
where $\operatorname{sim}(\cdot,\cdot)$ denotes the similarity measure used by the retrieval engine.

The retrieved neighbors provide historical references whose recent sensor behavior is close to that of the test unit. Their metadata are used to recover the corresponding training trajectories, cycle positions, and RUL values. These references are then used in the next stage to generate the visual comparison artifact for multimodal prompting.

\subsection{Multimodal Prompt Generation and MLLM Inference}
\label{sec:mm-fusion}

The multimodal prompt-generation stage is composed of three main modules: the trajectory comparison builder, the multimodal prompt composer, and the structured output parser. Figure~\ref{fig:mm_prompt_generation} summarizes the overall process, showing how the query trajectory and the retrieved historical references are transformed into a visual comparison artifact, inserted into a structured multimodal prompt, and processed by the MLLM to produce an RUL estimate and a textual explanation.

\subsubsection{Trajectory Comparison Builder}
\label{sec:trajectory_comparison_builder}

The Trajectory Comparison Builder transforms the query trajectory and the retrieved historical references into a trajectory-comparison image. This image contrasts the recent behavior of the test unit with similar training trajectories, allowing the MLLM to inspect sensor levels, temporal trends, and degradation-stage alignment in a compact visual form.

As shown in Figure~\ref{fig:mm_prompt_workflow}, this visual artifact is generated before prompt composition and serves as the main grounding evidence for the subsequent MLLM-based RUL prediction.

\paragraph{Visual encoding choices.}

The visual artifact is designed to make the degradation comparison readable within the constraints of a paper figure and an MLLM input. The query trajectory and the retrieved references are plotted on a common temporal axis, so that their recent behavior can be compared directly. The selected sensor channels are shown as separate trajectories, allowing the model to evaluate whether the similarity between the query and the references is consistent across multiple variables.

The example reported in Figure~\ref{fig:mm_prompt_example} includes two selected sensors and the corresponding RUL profile for readability. In the actual inference pipeline, the comparison artifact incorporates the sensor variables selected during preprocessing, as described in Section~\ref{sec:preprocessing}. The RUL information associated with the retrieved training references is used to contextualize their degradation stage, while the RUL of the test unit remains the target to be estimated.

These visual encoding choices are intended to provide the MLLM with evidence about three main aspects: the current degradation level of the query unit, its recent temporal trend, and its similarity to historical run-to-failure examples.

\subsubsection{Multimodal Prompt Composer}
\label{sec:multimodal_prompt_composer}

The Multimodal Prompt Composer combines the trajectory-comparison image with a textual prompt template, as shown in Figure~\ref{fig:mm_prompt_workflow}. The resulting prompt provides the MLLM with visual evidence, a description of the RUL estimation task, instructions for comparing the query trajectory with the retrieved references, and an explicit output schema.

This structure ensures that all evaluated MLLMs receive a consistent multimodal input. The system prompt, instruction prompt, and user prompt template used in this study are reported in the Appendix.

\subsubsection{Structured Output Parsing}

For each query unit, the MLLM is required to return a structured response containing both the numerical RUL estimate and a short textual explanation. The expected output has the form
\[
y_i =
\left\{
\begin{array}{l}
\text{``rul''}: \hat{r}_i, \\
\text{``explanation''}: e_i
\end{array}
\right\},
\]
where $\hat{r}_i \in \mathbb{N}_0$ denotes the predicted RUL of the last observation of the query trajectory, and $e_i$ is the explanation generated by the model.

The predicted RUL value is used for quantitative evaluation, whereas the explanation provides qualitative information about the reasoning process followed by the MLLM. In this way, the model is used not only as a regressor, but also as a reasoning component that can justify its prediction with reference to the visible degradation evidence. A representative explanation returned by the model is provided in Section~\ref{sec:disc}.

\begin{figure*}
    \centering
    \includegraphics[width=0.8\linewidth]{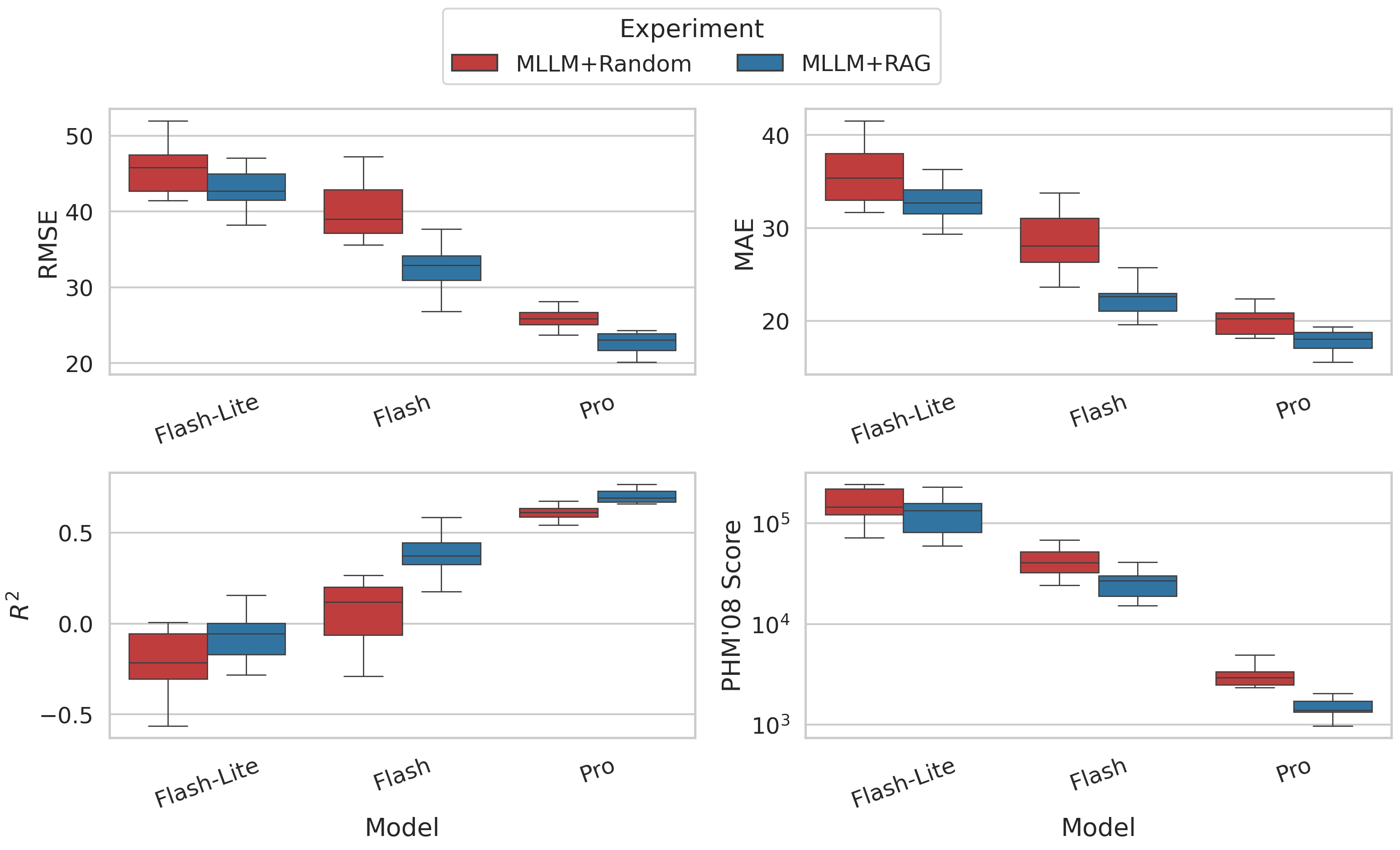}
    \caption{Predictive performance over 10 repetitions for the three evaluated Gemini MLLMs under the two inference settings, \textit{MLLM+Random} and \textit{MLLM+RAG}.}
    \label{fig:results}
\end{figure*}

\begin{table*}[t]
\centering
\caption{FD001 test-set comparison between baseline methods and Gemini-based RAG variants.}
\label{tab:fd001_baselines_vs_rag}
\renewcommand{\arraystretch}{1.15}
\footnotesize
\setlength{\tabcolsep}{4pt}
\begin{tabular}{lccc ccc}
\toprule
& \multicolumn{3}{c}{\textbf{Baselines}} & \multicolumn{3}{c}{\textbf{Gemini RAG}} \\
\cmidrule(lr){2-4} \cmidrule(lr){5-7}
\textbf{Metric} & \textbf{Mean} & \textbf{Random predictor} & \textbf{LSTM \cite{chenMachineRemainingUseful2021a}} & \textbf{Flash-Lite} & \textbf{Flash} & \textbf{Pro} \\
\midrule
RMSE
& 41.98 $\pm$ 0.00
& 57.19 $\pm$ 2.49
& 14.54 
& 43.12 $\pm$ 2.72
& 32.85 $\pm$ 3.30
& 22.69 $\pm$ 1.44 \\

MAE
& 36.07 $\pm$ 0.00
& 47.53 $\pm$ 2.66
& --
& 32.91 $\pm$ 2.19
& 22.47 $\pm$ 1.99
& 17.91 $\pm$ 1.24 \\

$R^2$
& -0.021 $\pm$ 0.000
& -0.897 $\pm$ 0.165
& --
& -0.080 $\pm$ 0.135
& 0.370 $\pm$ 0.125
& 0.701 $\pm$ 0.037 \\

PHM'08 Score
& 20302 $\pm$ 0
& 181531 $\pm$ 74903
& 322.44
& 127441 $\pm$ 66610
& 27712 $\pm$ 15830
& 1488 $\pm$ 323 \\
\bottomrule
\end{tabular}
\end{table*}

\subsection{Experimental Protocol}
\label{sec:experimental_protocol}

The experimental protocol compares two reference-selection strategies for multimodal RUL estimation: \emph{random} trajectory selection and \emph{RAG-based} trajectory selection. Let $\mathcal{M}$ denote the set of evaluated MLLMs. The experiment is repeated $N=10$ times to account for variability induced by the MLLM inference parameters and by the non-deterministic construction of the RAG component.

At each repetition $n$, a repetition-specific environment
\[
E^{(n)} = \bigl(\theta^{(n)}, \mathcal{C}^{(n)}, \mathcal{I}^{(n)}\bigr)
\]
is instantiated, where $\theta^{(n)}$ denotes the sampled inference parameters, $\mathcal{C}^{(n)}$ is the reference corpus derived from the training set, and $\mathcal{I}^{(n)}$ is the corresponding retrieval index. For each reference-selection strategy, the prompts are generated once under the current environment and then used to query all MLLMs in $\mathcal{M}$. This produces a paired evaluation setting in which, within each repetition, all models are assessed using the same prepared prompts.

\begin{algorithm}[t]
\SetAlgoLined
\KwIn{Train set $\mathcal{D}_{train}$, test set $\mathcal{D}_{test}$, window length $L$, number of references $k$, prompt template $T$, set of MLLMs $\mathcal{M}$, parameter search space $\Theta$, repetitions $N$}
\KwOut{Predictions and scores for all strategies, MLLMs, and repetitions}

\For{$n \leftarrow 1$ \KwTo $N$}{
    $E^{(n)} \leftarrow \textsc{BuildEnvironment}(\mathcal{D}_{train}, L, \Theta)$\;
    
    \ForEach{$s \in \{\texttt{random}, \texttt{rag}\}$}{
        $\mathcal{P}^{(n,s)} \leftarrow \textsc{Prompts}(\mathcal{D}_{test}, E^{(n)}, s, k, T)$\;
        
        \ForEach{$m \in \mathcal{M}$}{
            $\hat{\mathcal{Y}}^{(n,s,m)} \leftarrow \textsc{CallLLM}(\mathcal{P}^{(n,s)}, m, E^{(n)})$\;
            $M^{(n,s,m)} \leftarrow \textsc{Score}(\hat{\mathcal{Y}}^{(n,s,m)})$\;
        }
    }
}
\caption{Experimental protocol used to compare random and RAG-based reference selection across multiple MLLMs over repeated non-deterministic environments.}
\label{alg:experimental_protocol}
\end{algorithm}

Algorithm~\ref{alg:experimental_protocol} summarizes the full evaluation procedure. \textsc{BuildEnvironment} samples the inference parameters, constructs the reference corpus, and builds the retrieval index. \textsc{Prompts} extracts the last observed test segments, selects the references according to the considered strategy, generates the comparison figures, and instantiates the prompt template. Finally, \textsc{CallLLM} queries each MLLM, and \textsc{Score} evaluates the resulting RUL predictions. The scores are aggregated across repetitions to compare random and retrieval-based reference selection.

\paragraph{Evaluated MLLMs.}
The experimental evaluation considers three Gemini-based multimodal large language models, namely \textit{Gemini 3.1 Flash-Lite}, \textit{Gemini 3 Flash}, and \textit{Gemini 3.1 Pro}. These models were selected to cover different levels of model capacity and computational cost. In particular, Flash-Lite represents a lighter and more efficient variant, Flash provides an intermediate configuration, and Pro corresponds to a more capable model intended for more demanding reasoning tasks. This selection allows us to analyze whether the effect of retrieval augmentation is consistent across models with different capability--efficiency trade-offs.

\section{Results}
\label{sec:res}

Figure~\ref{fig:results} reports the predictive performance of the three evaluated Gemini MLLMs under the two considered inference settings, namely random multimodal inference (\textit{MLLM+Random}) and retrieval-augmented multimodal inference (\textit{MLLM+RAG}). Across all evaluated models, the inclusion of time-series retrieval leads to a consistent improvement in predictive quality, yielding lower error and more stable behavior than random reference selection. This trend indicates that the retrieved historical trajectories provide useful grounding evidence for the MLLM during RUL estimation.

Among the proposed configurations, the strongest results are obtained by the \textit{Pro + RAG} setting, while \textit{Flash + RAG} also shows clear gains over its non-retrieval counterpart. The \textit{Flash-Lite} variant benefits less from retrieval, suggesting that the usefulness of the retrieved information depends not only on the retrieval mechanism itself, but also on the reasoning capacity of the underlying multimodal model. Overall, the results support the claim that similarity-based reference selection is preferable to arbitrary reference sampling when MLLMs are used for RUL estimation.

The comparison reported in Table~\ref{tab:fd001_baselines_vs_rag} further helps position the proposed approach with respect to the literature. On the one hand, the proposed RAG-based variants outperform simple baselines and improve consistently over the corresponding random-reference MLLM setting. On the other hand, the best proposed configuration does not yet surpass the strongest task-specific and more sophisticated deep learning architecture included in the comparison. 

A second relevant observation concerns performance dispersion across the repeated experiments. For the \textit{Flash} and \textit{Pro} variants, the RAG-based setting tends to produce more compact distributions than the non-retrieval baseline, especially for the main error metrics. This suggests that retrieval not only improves the central tendency of the predictions, but also reduces sensitivity to arbitrary reference selection. In summary, the results indicate that time-series retrieval is a beneficial component within the proposed MLLM pipeline, even though the overall architecture remains simpler than the strongest predictive models reported in the literature.

\section{Discussion}
\label{sec:disc}

The experimental results provide a positive answer to the main research question of this paper: in the considered FD001 setting, time-series retrieval improves MLLM-based RUL estimation. The observed gains suggest that retrieved training trajectories provide useful grounding evidence, enabling the model to compare the query with historically similar degradation patterns rather than relying only on generic multimodal reasoning over the input artifact. In this sense, the proposed time-series RAG mechanism strengthens the link between the observed trajectory and previously seen run-to-failure behaviors.

Another relevant aspect concerns explainability. Unlike conventional regression baselines, the proposed framework can return not only a numerical RUL estimate but also a textual explanation of the rationale behind the prediction. Although this advantage does not compensate for the current performance gap with stronger specialized models, it remains valuable in PHM settings, where analysts may benefit from inspecting the prediction together with the retrieved historical references and the model's explanation. This explanatory capability is one of the motivations for exploring language-model-based approaches in prognostics.

To illustrate this qualitative aspect, Example~1 reports a representative explanation produced by the MLLM. The response shows that the model is able to relate the predicted RUL to observable properties of the query trajectory, such as the absence of visible degradation and the similarity with early-life train references. We also manually inspected a subset of the explanations and found them broadly consistent with the visual evidence provided in the trajectory-comparison figures. However, explanation quality was not evaluated systematically in this study.

\begin{quote}
\textbf{Example 1. Representative explanation returned by the MLLM.}

\small
The TEST unit's sensor readings across all informative channels (such as s2, s3, s4, s7, s11, and s12) are completely flat and show no signs of degradation. The levels closely match the baseline values of the TRAIN references during their earliest recorded cycles, well before any degradation trends begin. Since there is no visible progression towards failure, the unit is clearly in its early life stage. Therefore, the estimated RUL is 125 cycles.
\end{quote}

At the same time, the comparison with the literature must be interpreted carefully. The proposed approach does not yet match the predictive accuracy of the strongest architectures included in Table~\ref{tab:fd001_baselines_vs_rag}. However, this gap should be read in light of the purpose of the study. The main objective of this paper is not to introduce the most sophisticated possible multimodal architecture for RUL prediction, but rather to isolate and evaluate the influence of time-series RAG within a comparatively simple and controlled pipeline.

Several additional limitations should nevertheless be acknowledged. First, the evaluation is restricted to FD001, which is the simplest C-MAPSS subset. Therefore, the findings should not yet be generalized to more challenging settings involving multiple operating conditions, multiple fault modes, or previously unseen failure mechanisms. In particular, the proposed time-series RAG component can only retrieve degradation patterns that are represented in the indexed historical memory. If a failure mode has not occurred before, or is not sufficiently covered by the training trajectories, the system may retrieve only partially similar references and provide misleading grounding evidence to the MLLM. Second, the current study relies on a specific prompt design, a specific visual comparison artifact, and a single retrieval formulation, which may influence the quality of both the numerical predictions and the generated explanations. Third, although the experiment is repeated multiple times to account for variability, MLLM-based inference remains inherently non-deterministic and the proposed framework does not provide a formal guarantee of correctness. The repeated evaluation should therefore be interpreted as an empirical robustness assessment rather than as a correctness guarantee. Finally, the present analysis remains mainly descriptive and does not yet include systematic uncertainty estimation, novelty detection, or mechanisms for flagging cases in which the retrieved evidence is insufficient. Altogether, these limitations support interpreting the present work as a preliminary study on the role of retrieval in MLLM-based prognostics.

\section{Conclusions}
\label{sec:conc}

This paper presented a framework for studying RUL estimation with multimodal language models under retrieval augmentation. The proposed approach combines a time-series RAG module, which retrieves historically similar train segments, with a multimodal prompting pipeline that transforms the query and retrieved references into an input suitable for MLLM-based inference.

The empirical results on FD001 show that retrieval augmentation consistently improves the proposed MLLM pipeline relative to non-retrieval multimodal inference. Across all evaluated Gemini variants, the RAG-based setting yields better predictive performance and more stable behavior than random reference selection. These results support the main conclusion of the paper: time-series RAG has a positive influence on multimodal RUL estimation.

More broadly, the study highlights two promising aspects of language-model-based approaches for prognostics. First, they can benefit from explicit access to relevant historical trajectories rather than relying only on internal model reasoning. Second, they can naturally provide textual explanations alongside the numerical prediction, which may be valuable in human-centered PHM decision support settings.

Several directions can be explored in future work. First, the evaluation should be extended beyond FD001 to the more challenging C-MAPSS subsets and to additional predictive maintenance datasets. Second, the retrieval component could be improved through stronger embedding models, alternative similarity measures, and re-ranking strategies tailored to degradation trajectories. Third, future work should investigate how time-series RAG interacts with more advanced multimodal architectures, including richer temporal backbones and more effective cross-modal fusion mechanisms. Finally, it would be valuable to study jointly the relationship between retrieval quality, predictive accuracy, and explanation quality in order to better understand when MLLM-based prognostic reasoning is most useful in practice.

\section*{Acknowledgment}
\label{sec:acknowledgment}
This research was funded in whole, or in part, by the Luxembourg National Research Fund (FNR), grant reference \allowbreak BRIDGES/2022/IS/17270233. For the purpose of open access, and in fulfillment of the obligations arising from the grant agreement, the authors have applied a Creative Commons Attribution 4.0 International (CC BY 4.0) license to any Author Accepted Manuscript version arising from this submission.
\bibliographystyle{apacite}

@article{yinSurveyMultimodalLarge2024,
  author  = {Yin, Shukang and Fu, Chaoyou and Zhao, Sirui and Li, Ke and Sun, Xing and Xu, Tong and Chen, Enhong},
  title   = {A Survey on Multimodal Large Language Models},
  journal = {National Science Review},
  volume  = {11},
  number  = {12},
  pages   = {nwae403},
  year    = {2024},
  month   = nov,
  doi     = {10.1093/nsr/nwae403}
}

@article{jiangArtificialIntelligenceAgentEnabled2025,
  author  = {Jiang, Wenyu and Hu, Fuwen},
  title   = {Artificial Intelligence Agent-Enabled Predictive Maintenance: Conceptual Proposal and Basic Framework},
  journal = {Computers},
  volume  = {14},
  number  = {8},
  pages   = {329},
  year    = {2025},
  month   = aug,
  doi     = {10.3390/computers14080329}
}

@article{kirubanandanCausalAwareLLMAgents2025,
  author  = {Kirubanandan, Rajarajan},
  title   = {Causal-Aware {LLM} Agents for {PHM} Co-Pilots: Health Monitoring and Intervention Planning},
  journal = {Annual Conference of the PHM Society},
  volume  = {17},
  number  = {1},
  year    = {2025},
  doi     = {10.36001/phmconf.2025.v17i1.4321}
}

@inproceedings{saxenaDamagePropagationModeling2008c,
  author    = {Saxena, Abhinav and Goebel, Kai and Simon, Don and Eklund, Neil},
  title     = {Damage propagation modeling for aircraft engine run-to-failure simulation},
  booktitle = {2008 International Conference on Prognostics and Health Management},
  address   = {Denver, CO, USA},
  publisher = {IEEE},
  pages     = {1--9},
  year      = {2008},
  month     = oct,
  doi       = {10.1109/PHM.2008.4711414}
}

@inproceedings{decastroDeanFrederickSaratoga2007,
  author    = {DeCastro, Jonathan and Litt, Jonathan and Frederick, Dean},
  title     = {A Modular Aero-Propulsion System Simulation of a Large Commercial Aircraft Engine},
  booktitle = {44th {AIAA}/{ASME}/{SAE}/{ASEE} Joint Propulsion Conference \& Exhibit},
  address   = {Hartford, Connecticut},
  publisher = {American Institute of Aeronautics and Astronautics},
  number    = {AIAA 2008-4579},
  year      = {2008},
  doi       = {10.2514/6.2008-4579}
}

@misc{zhangLargeLanguageModels2024,
  author        = {Zhang, Xiyuan and Chowdhury, Ranak Roy and Gupta, Rajesh K. and Shang, Jingbo},
  title         = {Large Language Models for Time Series: A Survey},
  publisher     = {arXiv},
  year          = {2024},
  month         = may,
  doi           = {10.48550/arXiv.2402.01801},
  archivePrefix = {arXiv},
  eprint        = {2402.01801},
  primaryClass  = {cs.LG}
}

@article{chenMachineRemainingUseful2021a,
  author  = {Chen, Zhenghua and Wu, Min and Zhao, Rui and Guretno, Feri and Yan, Ruqiang and Li, Xiaoli},
  title   = {Machine Remaining Useful Life Prediction via an Attention-Based Deep Learning Approach},
  journal = {IEEE Transactions on Industrial Electronics},
  volume  = {68},
  number  = {3},
  pages   = {2521--2531},
  year    = {2021},
  month   = mar,
  doi     = {10.1109/TIE.2020.2972443}
}

@article{renPHMGPTLarge2025,
  author  = {Ren, Jiaxin and Liu, Xue and Wang, Tianlei and Zhao, Zhibin and Chen, Xuefeng and Li, Weihua and Yan, Ruqiang},
  title   = {{PHM-GPT}: A Large Language Model for Prognostics and Health Management},
  journal = {Engineering},
  pages   = {S2095809925006745},
  year    = {2025},
  month   = nov,
  doi     = {10.1016/j.eng.2025.11.001}
}

@article{hafsiPotentialGenerativeArtificial2025,
  author  = {Hafsi, Meriem},
  title   = {Potential of Generative {AI} in Knowledge-Based Predictive Maintenance for Aircraft Engines},
  journal = {Annual Conference of the PHM Society},
  volume  = {17},
  number  = {1},
  year    = {2025},
  doi     = {10.36001/phmconf.2025.v17i1.4352}
}

@inproceedings{tanPRETRAINEDLLMBASEDREMAINING2025,
  author = {Tan, Qingcheng and Yang, Lechang and Zhu, Feng and Wang, Zhe},
  title = {Pre-trained {LLM}-based remaining useful life prediction of aircraft engines},
  booktitle = {2025 15th International Conference on Quality, Reliability, Risk, Maintenance, and Safety Engineering ({QR2MSE}) and 8th International Conference on Materials and Reliability ({ICMR})},
  series = {IET Conference Proceedings},
  volume = {2025},
  number = {35},
  pages = {1016--1024},
  year = {2025},
  doi = {10.1049/icp.2025.3534}
}

@misc{chenRemainingUsefulLife,
  author        = {Chen, Yan and Liu, Cheng},
  title         = {Remaining Useful Life Prediction: A Study on Multidimensional Industrial Signal Processing and Efficient Transfer Learning Based on Large Language Models},
  publisher     = {arXiv},
  year          = {2024},
  month         = oct,
  doi           = {10.48550/arXiv.2410.03134},
  archivePrefix = {arXiv},
  eprint        = {2410.03134},
  primaryClass  = {cs.LG}
}

@inproceedings{lewisRetrievalAugmentedGenerationKnowledgeIntensive2021,
  author = {Lewis, Patrick and Perez, Ethan and Piktus, Aleksandra and Petroni, Fabio and Karpukhin, Vladimir and Goyal, Naman and K{"u}ttler, Heinrich and Lewis, Mike and Yih, Wen-tau and Rockt{"a}schel, Tim and Riedel, Sebastian and Kiela, Douwe},
  title = {Retrieval-Augmented Generation for Knowledge-Intensive {NLP} Tasks},
  booktitle = {Advances in Neural Information Processing Systems},
  volume = {33},
  pages = {9459--9474},
  year = {2020},
  publisher = {Curran Associates, Inc.},
  archivePrefix = {arXiv},
  eprint = {2005.11401},
  primaryClass = {cs.CL},
  doi = {10.48550/arXiv.2005.11401}
}

@misc{wangTSMLLMMultiModalLarge2026,
  author        = {Wang, Haiteng and Li, Yikang and Zhu, Yunfei and Yan, Jingheng and Ren, Lei and Yang, Laurence T.},
  title         = {{TS-MLLM}: A Multi-Modal Large Language Model-based Framework for Industrial Time-Series Big Data Analysis},
  publisher     = {arXiv},
  year          = {2026},
  month         = mar,
  doi           = {10.48550/arXiv.2603.07572},
  archivePrefix = {arXiv},
  eprint        = {2603.07572},
  primaryClass  = {cs.LG}
}
\PHMbibliography{bibliography}

\newpage

\begin{strip}

\section*{Appendix}
\label{sec:appendix}
\subsection*{Prompt Templates}

\begin{center}

\begin{minipage}{0.95\textwidth}
\begin{promptbox}{\faRobot\quad System Prompt}
\footnotesize
You are an expert Remaining Useful Life (RUL) estimation analyst for industrial equipment monitored over time through multiple sensors.

\vspace{0.35em}
Your task is to estimate the RUL of the TEST unit at the most recent cycle shown.

\vspace{0.35em}
You will receive:
\begin{itemize}[leftmargin=*, itemsep=1pt, topsep=2pt, parsep=0pt]
  \item a plot comparing a TEST time series against one or more TRAIN time series;
  \item optional contextual text describing how to interpret the trajectories.
\end{itemize}

\vspace{0.35em}
\textbf{Decision policy:}
\begin{enumerate}[leftmargin=*, itemsep=1pt, topsep=2pt, parsep=0pt]
  \item Estimate the TEST unit's latest RUL directly from the visual evidence.
  \item Compare the TEST unit's recent sensor behavior against the TRAIN references.
  \item Use visible evidence such as level, slope, curvature, and degradation stage to judge where the TEST unit lies in its degradation process.
  \item Infer the RUL by assessing which TRAIN degradation stage the TEST unit most closely resembles, with emphasis on the most recent cycles.
  \item Use agreement across multiple sensors more than any single noisy signal.
  \item If the evidence is weak, noisy, or contradictory, return a cautious estimate consistent with the overall visible trend.
\end{enumerate}

\vspace{0.35em}
\textbf{Constraints:}
\begin{itemize}[leftmargin=*, itemsep=1pt, topsep=2pt, parsep=0pt]
  \item Output MUST be valid JSON matching the schema.
  \item \texttt{rul} must be an integer in \texttt{[0,\{clip\_rul\}]}.
  \item \texttt{explanation} must contain 2--5 sentences and must be grounded only in visible trends from the plot.
\end{itemize}
\end{promptbox}
\end{minipage}

\vspace{0.7em}

\begin{minipage}{0.95\textwidth}
\begin{promptbox}{Instructions}
\footnotesize
\begin{itemize}[leftmargin=*, itemsep=1pt, topsep=2pt, parsep=0pt]
  \item Focus on recent degradation trend and relative alignment to train trajectories.
  \item Give more weight to multi-sensor agreement than to isolated fluctuations.
  \item Infer degradation stage from visible evidence, not from any assumed external prediction.
  \item Prefer estimates that are visually well-supported by the TEST-to-TRAIN comparison.
  \item Keep uncertainty calibrated when the evidence is weak.
\end{itemize}
\end{promptbox}
\end{minipage}

\vspace{0.7em}

\begin{minipage}{0.95\textwidth}
\begin{promptbox}{\faRobot\quad User Prompt Template}
\footnotesize
Task: estimate the Remaining Useful Life (RUL) of the TEST time series shown in this plot.

\vspace{0.35em}
\textbf{Problem description:}
\begin{itemize}[leftmargin=*, itemsep=1pt, topsep=2pt, parsep=0pt]
  \item Each unit is monitored over time through multiple sensor measurements.
  \item The TRAIN references show complete run-to-failure trajectories or long degradation trajectories from similar units.
  \item The TEST unit is only observed up to its current cycle, and your goal is to estimate how many cycles of useful life remain.
  \item Lower RUL means the unit appears closer to failure or to a late degradation stage.
  \item Higher RUL means the unit appears to be in an earlier degradation stage.
\end{itemize}

\vspace{0.35em}
\textbf{Important:}
\begin{itemize}[leftmargin=*, itemsep=1pt, topsep=2pt, parsep=0pt]
  \item Estimate the RUL directly from the plot.
  \item Do not assume any external model prediction is available.
  \item Base your estimate only on the visible sensor behavior and alignment with the TRAIN references.
\end{itemize}

\vspace{0.35em}
\textbf{What to do:}
\begin{itemize}[leftmargin=*, itemsep=1pt, topsep=2pt, parsep=0pt]
  \item Compare the TEST trajectory against the TRAIN references, especially in the most recent cycles.
  \item Focus on the recent degradation stage of the TEST unit.
  \item Use agreement across multiple sensors more than any single noisy signal.
  \item If the TEST unit appears close to late-life degradation patterns, predict a lower RUL.
  \item If the TEST unit appears to be at an earlier degradation stage, predict a higher RUL.
  \item If the evidence is mixed, provide a cautious estimate that best matches the overall visual pattern.
\end{itemize}

\vspace{0.35em}
\textbf{Return:}
\begin{itemize}[leftmargin=*, itemsep=1pt, topsep=2pt, parsep=0pt]
  \item \texttt{rul}: the estimated integer RUL for the most recent TEST cycle;
  \item \texttt{explanation}: a short explanation of how the visible sensor evidence supports the estimate.
\end{itemize}

\vspace{0.35em}
Before generating the final answer, ensure that the predicted RUL value is compliant with all constraints and aligned with the explanation.
\end{promptbox}
\end{minipage}

\vspace{0.7em}

\captionof{figure}{Prompt templates used for visual RUL estimation.}
\label{fig:prompt_templates}

\end{center}

\end{strip}

\end{document}